# Parsing in OT[*]

Michael Hammond
University of Arizona

## 0. Abstract

In this paper, I propose to implement syllabification in OT as a parser. I propose several innovations that result in a finite and small candidate set: i) MAX and DEP violations are not hypothesized by the parser, ii) candidates are encoded locally, and iii) EVAL is applied constraint by constraint. The parser is implemented in Prolog, and has a number of desirable consequences. First, it runs and thus provides an existence proof that syllabification can be implemented in OT. Second, GEN is implemented as a finite-state transducer. Third, the implementation makes an exponential savings in processing time over previous approaches. Fourth, the implementation suggests some particular reformulations of some of the benchmark constraints in the OT arsenal, e.g. *COMPLEX, PARSE, ONSET, and NOCODA. Finally, the implementation is compared with various other proposals in the literature.

## 1. Overview

In this paper I argue that Optimality Theory (OT) provides for an explanatory model of phonological parsing. I show how on first blush there seem to be insurmountable difficulties in implementing OT as a parser. On closer inspection, relaxing three key aspects of OT enables us to implement an efficient parser with a number of interesting properties. First, MAX and DEP violations are not entertained by the parser. Second, the candidate set is encoded locally. Third, CON and EVAL are applied cyclically.

The organization of this paper is as follows. First, I outline how OT works, and how it describes syllabification. I then show how the most simple-minded parser built on these assumptions fails. The following sections make a number of incremental improvements to the parser which result finally in a working and highly efficient implementation. This parser makes predictions about linguistic structure which are then reviewed and shown to be correct. Finally, alternative implementations of OT as a parser are then considered.

## 2. Syllables

Syllabification is a hotly debated and active research area in current phonological theory.[1] The basic idea is that all languages exhaustively parse a string of segments into syllables, where each syllable can be defined as a single peak of sonority or intrinsic loudness. For example, a word like

---

[*] Thanks to Sonya Bird, Chris Golston, Young-shik Hwangbo, Diane Ohala, Markus Walther, Richard Wiese, and the participants in my fall '97 seminar on computational phonology for much useful discussion. This paper describes a parser implemented in Prolog. The code is available from the author and via the web (http://www.u.arizona.edu/~hammond). The parser can also be run over the web at the same URL. All errors are the author's.

[1] For more discussion of syllabification, see Hammond (1997b) and the references cited therein.

*a.gen.da* would contain three syllables (separated by periods). A word like *cramp*, on the other hand, contains only one syllable, where the vowel constitutes the peak and sonority falls toward the edges of the syllable.

Languages differ widely in terms of the syllables allowed. There are a number of parameters of variation. For example, what can constitute a sonority peak can vary. English allows any vowel and a few other sounds [l,r,n,m] to function as peaks. For example, the word *table* has two peaks where the second peak is [l]. Spanish, on the other hand, does not allow anything but a vowel to be a syllable peak.

Languages also vary in what kinds of material is allowed on the edges of the syllable (the onset and coda respectively). Some languages, like English, allow consonants to appear or not at either edge, e.g. *toe*, *oat*, *tote*, *oh*.[2] Some languages, like Hawaiian, disallow any consonants on the right; while others require a consonant in the syllable onset, e.g. the language Senufo. There are four cases and they can be schematized.

(1)  English        (C)V(C)
     Yawelmani      CV(C)
     Hawaiian       (C)V
     Senufo         CV

Strikingly, other imaginable cases do not occur.[3] For example, there are no languages where a consonant is required on the right side of the syllable: *(C)VC.

A third dimension along which syllables can vary is how many segments can occur as an onset or coda. In English, multi-segment onsets and codas are permitted; in Yawelmani, they are not.

There is lots more to say about syllables, but this will suffice to exemplify how Optimality Theory can be applied to syllabification.

## 3.  Optimality Theory

Prince & Smolensky (1993) propose that syllabification and other phonological generalizations should be treated in terms of constraint hierarchies rather than in terms of a sequence of rules. The basic idea is that every possible syllabification of an unsyllabified input string is generated by the function GEN and then evaluated by the function EVAL with respect to the constraint set CON. Extremely important aspects of this system are that the constraints are violable and strictly ranked. Violability means that even the best candidate may violate some constraints. Strict ranking means that the constraints are ordered such that a single violation of a higher-ranked constraint is worse than any number of violations of a lower-ranked constraint.

These points can be seen in the following examples. First, let's consider an abstract case. Imagine that GEN produces two candidate syllabifications for some input string. Imagine that CON consists of two constraints ranked in the order given (2) and that violations are as given.

---

[2] See Hammond (1987a, to appear) for discussion of the intricacies of English syllabification.
[3] This observation is due to Roman Jakobson. See Jakobson (1962).

(2)

| /input/ | constraint 1 | constraint 2 |
|---|---|---|
| ☞ candidate a |  | * |
| candidate b | *! |  |

In this tableau, candidate a is judged optimal by EVAL. Only candidate b violates constraint 1, and therefore candidate a is superior. The fact that candidate a violates constraint 2 is irrelevant because constraint 2 is lower ranked than constraint 1 and the violation of constraint 1 cannot be outweighed by the violation of constraint 2, or any number of possible violations of constraint 2. This ranking of the violations is indicated in (2) with the exclamation point and shading of areas of the tableau that are then irrelevant.

Let's consider a more linguistic example. Here I adopt two constraints from the literature – DEP (3) and ONSET (4). DEP requires that all segments be underlying, and thus limits epenthesis of vowels or consonants.[4]

(3)  DEP
     All segments must be underlying.

The ONSET constraint requires that all syllables have onsets.

(4)  ONSET
     Syllables must have onsets.

Consider now how an input string like *Aggie* /ægi/ is syllabified. Syllable boundaries are marked with periods and epenthetic segments with outline.

(5)

| /ægi/ | DEP | ONSET |
|---|---|---|
| ☞ a  æ.gi |  | * |
| b  tæ.gi | *! |  |
| c  æg.i |  | **! |
| d  tæg.i | *! | * |

This tableau illustrates several points. First, candidates where ONSET is satisfied by an epenthetic consonant are thrown out because of high-ranking DEP. Second, note how the optimal candidate violates ONSET. Third, notice how candidate (5c) is ruled out because it violates ONSET twice, while the winning candidate only violates it once.

Notice that another way of avoiding an ONSET violation would be to leave out the vowel [æ]. This strategy is mediated by the constraint MAX which says that all input segments must be pronounced.

(6)  MAX
     All input segments must be pronounced.

---

[4] The DEP constraint is drawn from a revision to OT called Correspondence Theory (McCarthy & Prince, 1995).

When these candidates are added and MAX is considered as well, tableau (5) is expanded to (7).

(7)

|   | /ægi/ | MAX | DEP | ONSET |
|---|---|---|---|---|
| ☞ a | æ.gi | | | * |
| b | tæ.gi | | *! | |
| c | æg.i | | | **! |
| d | tæg.i | | *! | * |
| e | gi | *! | | |

The ranking of MAX and DEP cannot be determined, but MAX must outrank ONSET. In a language like Senufo where onsets are required, the ONSET would be ranked higher.

Constraints like MAX and DEP are called faithfulness constraints because they regulate the mapping between two different representations. Constraints like ONSET are called well-formedness constraints because they govern only the well-formedness of some particular representation.

Whether coda consonants are allowed is also handled by constraint. The NOCODA constraint disallows coda consonants.

(8)  NOCODA
Syllables can't have codas.

Consider now a word like *beet* [bit]. As with the ONSET constraint, to insure that this word surfaces with its coda consonant intact, MAX and DEP must outrank NOCODA.

(9)

|   | /bit/ | MAX | DEP | NOCODA |
|---|---|---|---|---|
| ☞ a | bit | | | * |
| b | biti | | *! | |
| c | bi | *! | | |

As in the previous example, the relative ranking of MAX and DEP cannot be determined on these data. A language like Hawaiian where codas are disallowed would rank NOCODA higher.

This theory offers a straightforward account of the typological generalization made in (1) above. If we disregard the difference between MAX and DEP, grouping them together as FAITH, we have six logically possible rankings of FAITH, ONSET, and NOCODA.

(10) a. FAITH >> ONSET >> NOCODA
    b. FAITH >> NOCODA >> ONSET
    c. ONSET >> FAITH >> NOCODA
    d. NOCODA >> FAITH >> ONSET
    e. ONSET >> NOCODA >> FAITH
    f. NOCODA >> ONSET >> FAITH

Given the form of the ONSET and NOCODA constraints, it turns out that any rankings where those two constraints are adjacent is nondistinct. This reduces six to only four distinct rankings.

(11) a. FAITH >> $\begin{Bmatrix} \text{NOCODA} \\ \text{ONSET} \end{Bmatrix}$
 b. ONSET >> FAITH >> NOCODA
 c. NOCODA >> FAITH >> ONSET
 d. $\begin{Bmatrix} \text{NOCODA} \\ \text{ONSET} \end{Bmatrix}$ >> FAITH

Each of these rankings can be uniquely associated with one of the types in (1). Under ranking (11a) the pressure to pronounce all and only the segments of the input outweighs the pressure for an onset and against having a coda, resulting in the syllabic pattern of English. Under ranking (11b), the pressure to pronounce things as they are only outranks the NOCODA constraint. This means that onsets are required and codas are optional, e.g. Yawelmani. Ranking (11c) is just the opposite: onsets are optional and codas are strictly disallowed, e.g. Hawaiian. Finally, under ranking (11d), the well-formedness constraints both outrank FAITH resulting in the Senufo pattern.

Several more constraints are necessary to account for the variation in syllable type discussed above. First, *COMPLEX prohibits syllable margins containing more than one segment.

(12) *COMPLEX
 Syllable margins cannot contain more than one segment.

Lastly, the following two constraints prohibit consonants as peaks and vowels as margins.

(13) *PEAK/C
 A consonant cannot be a syllable peak.

(14) *MARGIN/V
 A vowel cannot be an onset or a coda.

These constraints will suffice to demonstrate the problems OT poses for a model of parsing.[5]

## 4. The problems

There are several problems facing an implementation of OT for syllabification. These are discussed in this section.

There are at least two ways in which one might think of implementing OT: as a generator or as a parser. A generator would map input forms onto properly syllabified output forms. A parser would take output forms and map them onto input forms. Since it is not at all clear that raw syllabification information is present in the phonetic form (though it is surely deducible), even a parser would have to supply syllable structure to an output string.

---

[5] For more discussion of how to apply OT to syllabification, see Hammond (1997b).

This means that there is a common element to both of these tasks. The generator would start with an input form, generate candidate syllabifications, and apply constraints to produce a syllabified output. A parser would start with an unsyllabified output, generate candidates, and produce a syllabified output.

The difference between these two kinds of implementations shows up in the treatment of epenthesis and deletion. The generator's task entails that any potential input segment could be deleted (violating MAX) or that epenthesis could occur in any position (violating DEP). The parser, on the other hand, is not responsible for hypothesizing epenthesis or deletion; it simply provides for the best syllabification of the segments it's faced with. The task of figuring out which segments are epenthetic and which segments are deleted can be left up to a separate lookup module.

This distinction in whether possible deletions or epenthesis sites are correctly guessed is a reasonable one. It's reasonable because we know speakers of a language can judge the syllabic well-formedness of nonsense words. Presumably this is done on the basis of a successful parse using the constraints and ranking of their language. The fact that the word cannot be found in their mental lexicon doesn't impinge on the well-formedness of such words. For example, speakers of English will judge a word like *blick* as well-formed, but a word like *bnick* as ill-formed, even though neither is a word of English.

It turns out that hypothesizing epenthesis and deletion sites is an extremely complex problem and so it is worthwhile to start work on implementing a parser, since that is a much more tractable domain.

The principal problem facing any implementation of OT as a generator that hypothesizes deletion/epenthesis sites is that the candidate set is infinite. There is no upper bound on the number of epenthetic elements a string may contain. Consider a hypothetical input form *beet* /bit/. Epenthetic elements can go between any of the segments or at either edge of the span. (Possible epenthesis sites are marked with underscores.)

(15)  _b_i_t_

Even if we limit the class of available epenthetic elements, the problem remains. The problem is that there can be an unbounded number of epenthetic elements between each segment. Prince & Smolensky analyze Axininca Campa as requiring just this.

Even if we limit epenthesis in each position to a single segment, there is still an explosion in the candidate set. For an input with $n$ segments, there are $2^{n+1}$ possible candidates to consider, just restricting ourselves to epenthesis. For the input /bit/ and an epenthetic element /a/, the following candidates need to be considered ($2^{3+1}=16$).

(16) | bita | baita | abita | abaita
     | biat | baiat | abiat | abaiat
     | biata | baiata | abiata | abaiata
     | bait | abit | abait | bit

Phonological deletion poses a similar problem. Assuming an input of *n* segments and assuming every combination of segments is deletable (up to deleting the entire input), there are $(2^n)-1$ candidates to consider. Here are the candidates for /bit/

(17) | bit | it
     | bi  | i
     | bt  | t
     | b   |

When one puts these two together and adds in the candidates generated from other options, the problem becomes intractable. The following chart shows different numbers for just these two problems for inputs of different lengths (given the assumptions made so far).

(18)

| segments | epenthesis | deletion | both |
|----------|------------|----------|------|
| 1 | 4 | 1 | 4 |
| 2 | 8 | 3 | 16 |
| 3 | 16 | 7 | 52 |
| 4 | 32 | 15 | 160 |
| 5 | 64 | 31 | 484 |
| 6 | 128 | 63 | 1456 |
| 7 | 256 | 127 | 4372 |

The answer is to separate the problems. The proposal is that hypothesis of epenthesis and deletion is separate from the other options available in the candidate set. Put another way in the parsing domain, first, the parser finds the best syllabification of the string; then the lookup machine undertakes to hypothesize epenthesis and deletion sites. This has three desirable consequences. First, it removes epenthesis and deletion as factors in calculating the size of the candidate set. Second, it allows the lookup machine to make use of the lexicon in bounding its search space. Third, it allows us to model the judgment of syllabic well-formedness.

Once we leave aside epenthesis and deletion (violations of MAX and DEP), the candidate set is clearly finite. However, it still remains intractably large. There are two principal reasons for this.

The first problem is exponential and therefore graver. Assume that any segment can in principle be syllabified as an onset, coda, or peak (nucleus). A fourth possibility is that the parser fails to syllabify the relevant element. This means that a word with *n* segments has $4^n$ possible syllabifications. The math is again frightening and diagrammed below.

*Syllable parsing/p.7*

(19)

| segments | candidates |
|---|---|
| 1 | 4 |
| 2 | 16 |
| 3 | 64 |
| 4 | 256 |
| 5 | 1,024 |
| 6 | 4,096 |
| 7 | 16,384 |
| 8 | 65,536 |
| 9 | 262,144 |
| 10 | 1,048,576 |

The second problem is arithmetic, but is still nontrivial. The problem is that each candidate has to be considered by each constraint. Thus if there are 100 candidates and 10 constraints, each candidate will be considered 10 times, giving 1000 combinations.

If we add this to the problem above, we get the following clear numerology. The following chart indicates the interaction of candidate set size and number of constraints. The size of the candidate set is a function of the number of segments and this is indicated along the left side. The number of constraints is indicated along the top.

(20)

|  | 1 | 2 | 3 | 4 | 5 |
|---|---|---|---|---|---|
| 1 | 4 | 8 | 16 | 32 | 64 |
| 2 | 16 | 32 | 64 | 128 | 256 |
| 3 | 64 | 128 | 256 | 512 | 1,024 |
| 4 | 256 | 512 | 1,024 | 2,048 | 4,096 |
| 5 | 1,024 | 2,048 | 4,096 | 8,192 | 16,384 |
| 6 | 4,096 | 8,192 | 16,384 | 32,768 | 65,536 |
| 7 | 16,384 | 32,768 | 65,536 | 131,072 | 262,144 |
| 8 | 65,536 | 131,072 | 262,144 | 524,288 | 1,048,576 |
| 9 | 262,144 | 524,288 | 1,048,576 | 2,097,152 | 4,194,304 |
| 10 | 1,048,576 | 2,097,152 | 4,194,304 | 8,388,608 | 16,777,216 |

In the following section, I outline a solution to both of these problems.

## 5. Parsing

The basic idea to be explored here is threefold. First, the parser will <u>not</u> analyze epenthetic elements or hypothesize unparsed segments and thus, since the number of attachment points at any point is finite, the number of candidates considered will be finite. Second, the model here makes use of a cyclic CON-EVAL loop to reduce the number of times multiple constraints consider the same candidate. Third, the model here makes use of local encoding of syllabification to reduce the candidate set dramatically.

Let us now consider these properties one by one.

First, the candidate set is always finite. Recall that this is because the parser does not hypothesize epenthetic segments or unparsed segments. This is clearly a contributor to resolving an optimal parse in finite time.

Second, this model makes use of a cyclic CON-EVAL loop. What this means is that violations are computed and assessed from the highest-ranked constraint down. Once a high-ranked constraint eliminates a candidate from consideration, violations of lower-ranked constraints by that candidate aren't even computed. In terms of the notation of the field, shaded violations aren't calculated. This can be seen in the following hypothetical tableau.

| (21) | A | B | C | D | E |
|---|---|---|---|---|---|
| 1 | *! | * |  |  | * |
| 2 | *! | * |  | * |  |
| 3 | *! |  |  |  | * |
| 4 | *! |  |  | * |  |
| 5 | *! | * |  |  |  |
| 6 |  | *! |  |  | * |
| 7 |  |  |  | * |  |
| 8 |  |  |  | * | *! |
| 9 |  | *! |  |  |  |
| 10 |  | *! |  |  |  |

The assumption standardly made is that violations for all constraints for all candidates are calculated. However, logically irrelevant violations need not ever be calculated. Once some constraint renders a candidate out of the running, violations of lower-ranked constraints by those candidates are irrelevant and can be skipped. What this means, for example, in terms of the tableau above is that instead of computing 50 constraint-candidate combinations, only 21 need be calculated.

The savings that result depend on the efficiency of the higher-ranked constraints. For example, if, in the tableau above, constraints A through D didn't throw out any candidates and all the work was done by constraint E, then there would be no savings. Thus 50 combinations out of 50 are considered in (22). On the other hand, if constraint A did all the work, there would be maximal savings. Out of 50 combinations, 10 are computed in (23). Both of these are diagrammed below.

(22)

| | A | B | C | D | E |
|---|---|---|---|---|---|
| 1 | * | | * | | *! |
| 2 | * | | * | | *! |
| 3 | * | | * | | *! |
| 4 | * | | * | | *! |
| 5 | * | | * | | *! |
| 6 | * | | * | | *! |
| 7 | * | | * | | *! |
| 8 | * | | * | | *! |
| 9 | * | | * | | *! |
| 10 | * | | * | | |

(23)

| | A | B | C | D | E |
|---|---|---|---|---|---|
| 1 | *! | | * | * | |
| 2 | *! | * | | * | |
| 3 | *! | * * | * | * | |
| 4 | *! | * * * | | * | |
| 5 | *! | | * | * | |
| 6 | *! | * | | * | |
| 7 | *! | * * | * | * | |
| 8 | *! | * * * | | * | |
| 9 | *! | | * | * | |
| 10 | | * | | * | |

If we assume that all rankings for these five constraints occur, then even if such differences occur, they will even out on average such that the savings across all grammars built on these constraints will be that approximately 30 comparisons will need to be made for a combination of this sort.

The math works as follows. Assume that all constraints are equally efficient. (If they are not, then their effect will be evened out under different rankings.) Thus if there are *n* constraints and *m* candidates, each constraint will rule out approximately *m/n* candidates. (Since, by the logic of OT, at least one candidate must remain, there will be some slippage on this figure.) The general result is diagrammed below.

(24)

| | A | B | C | D | E |
|---|---|---|---|---|---|
| 1 | *! | | | | |
| 2 | *! | | | | |
| 3 | | *! | | | |
| 4 | | *! | | | |
| 5 | | | *! | | |
| 6 | | | *! | | |
| 7 | | | | *! | |
| 8 | | | | *! | |
| 9 | | | | | *! |
| 10 | | | | | |

On this reasoning, cyclic application of EVAL after each constraint will make a rather substantive savings.

In fact, there is a second way to think of this which would result in more dramatic consequences. If we assume that, on average each constraint assigns violations to roughly half the candidates it faces, then we get something like the following.

(25)

| | A | B | C | D | E |
|---|---|---|---|---|---|
| 1 | *! | * | * | | |
| 2 | *! | | * | * | |
| 3 | *! | * | | * | * |
| 4 | *! | | | | |
| 5 | *! | * | * | | |
| 6 | | | *! | | * |
| 7 | | *! | | * | |
| 8 | | | | * | |
| 9 | | *! | * | | * |
| 10 | | | *! | | |

Here each constraint eliminates 1/2 of the remaining candidates, which means, in the present case, either 17 or 18 candidates are considered. Again, we might expect this to vary across constraints, but assuming all rankings occur and occur with equal frequency, an even more dramatic savings results crossgrammatically.

Let's now turn to the final proposal: local encoding. This "local coding" of the constraint set effectively means that even fewer candidates are considered in toto. The basic idea is to consider the structural options for each element separately. The comparison can be made as follows. Assume that any segment can be syllabified in any of four ways: onset (o), nucleus (n), coda (c), and unsyllabified (u). In a word with *n* segments, this means there are $4^n$ candidates. The string [pa], for example can be syllabified in 16 ways.

(26) oo    no    co    uo
     on    nn    cn    un
     oc    nc    cc    uc
     ou    nu    cu    uu

The idea behind local coding is the syllabification possibilities of each segment are considered separately; they are summed, rather than multiplied.

(27) {oncu} + {oncu}

I'll show in detail how this works in the following section, but if this general approach can work, it results in an exponential reduction in the candidate set. While orthodox nonlocal coding would result in $4^n$ candidates for an input with *n* segments, local coding results in 4*n* candidates.

(28)

| segments | local | nonlocal |
|---|---|---|
| 1 | 4 | 4 |
| 2 | 8 | 16 |
| 3 | 12 | 64 |
| 4 | 16 | 256 |
| 5 | 20 | 1,024 |
| 6 | 24 | 4,096 |
| 7 | 28 | 16,384 |
| 8 | 32 | 65,536 |
| 9 | 36 | 262,144 |
| 10 | 40 | 1,048,576 |

Thus, the implementation proposed drastically reduces the number of candidates that must be considered. It does so by i) generating only a finite number of candidate parses, ii) eliminating candidates in a cyclic fashion, and ii) coding candidate parses locally.

## 6.  The Implementation

The model described above is implemented in Prolog and makes use of several innovations that result in a relatively efficient resolution of the candidate set.[6] There are three main ones.

(29) a.   finite candidate set
     b.   serial constraint satisfaction (cyclic CON-EVAL)
     c.   local coding

In this section, I outline the program with particular attention to how the above are implemented. I'll show how it works with a particular input *agenda* with one ranking of the constraints currently implemented.

---

[6] This implementation makes use of the pure logical component of Prolog; the extralogical components are only used in IO.

The program first solicits an input item from the user, checks that it is composed of legal characters from the defined alphabet, and converts it to a Prolog list. For example, the input *agenda* would be converted to the following.

(30) [a,g,e,n,d,a]

The function GEN pairs this with a locally-encoded candidate set.[7] The internal representation is as follows.

(31) [a/[o,n,c,u],g/[o,n,c,u],e/[o,n,c,u],n/[o,n,c,u],d/[o,n,c,u],a/[o,n,c,u]]

Each symbol is associated with a list of the structural positions it can occupy, e.g. o:onset, n:nucleus, c:coda, u:unsyllabified.

I will recast this in a more perspicuous form as follows. I term this representation a candidate grid.

(32)
| a | g | e | n | d | a |
|---|---|---|---|---|---|
| o | o | o | o | o | o |
| n | n | n | n | n | n |
| c | c | c | c | c | c |
| u | u | u | u | u | u |

Each constraint then has the opportunity to prune away dispreferred candidates. This pruning operation can be seen as a paired CON-EVAL procedure where dispreferred candidate parses, those that violate the constraint in question, are first identified and then removed from the candidate set.

There is, for example, a constraint *PEAK/C which says that consonants prefer not to be syllable nuclei. This constraint removes the option 'n' from any segment identified as a consonant. If this constraint applied at the top, the candidate set for *agenda* would be reduced as follows.

(33)
| a | g | e | n | d | a |
|---|---|---|---|---|---|
| o | o | o | o | o | o |
| n |   | n |   |   | n |
| c | c | c | c | c | c |
| u | u | u | u | u | u |

Pruning is always subject to the condition that it does not reduce the candidate set to null.

(34) Pruning Condition
Pruning cannot reduce the candidate set to null.

---

[7] As shown below, this pairing can be seen as a finite state transducer.

The Pruning Condition thus enforces the idea that constraints in OT can be violated.

There is also a certain amount of housekeeping that needs to be done, given the particular encoding of syllable structure adopted. There are three principal configurations that are always resolved as they arise. First, a word cannot end in an onset. Second, a word cannot begin with a coda. Third, an onset cannot be followed by a coda. Before each pruning operation, housekeeping comes through to clean any of these up that arise.[8]

(35) Housekeeping
   The following configurations are removed by pruning
   whenever they arise.
   a. word-initial coda
   b. word-final onset
   c. an onset before a coda

In the example above, housekeeping will convert (33) to (36) below.

(36)
| a | g | e | n | d | a |
|---|---|---|---|---|---|
| o | o | o | o | o |   |
| n |   | n |   |   | n |
|   | c | c | c | c | c |
| u | u | u | u | u | u |

The *MARGIN/V constraint will prune 'o' and 'c' from any element identified as a vowel, producing the following.

(37)
| a | g | e | n | d | a |
|---|---|---|---|---|---|
|   | o |   | o | o |   |
| n |   | n |   |   | n |
|   | c |   | c | c |   |
| u | u | u | u | u | u |

The PARSE constraint eliminates 'u' whenever it can.

(38)
| a | g | e | n | d | a |
|---|---|---|---|---|---|
|   | o |   | o | o |   |
| n |   | n |   |   | n |
|   | c |   | c | c |   |

The ONSET constraint actually does several things. It penalizes any configuration where 'n' is not preceded by 'o'. It does so by eliminating any candidates it can. In the grid above, there are three cases where ONSET could in principle apply. First, ONSET can do nothing about the word-initial vowel. Epenthesis is not an option and the Pruning Convention precludes removing the 'n'. In the case of the second and third vowel, ONSET can prune the 'c' of the preceding consonants giving the following grid.

---

[8] Housekeeping is actually more general. The same configurations given are also ruled out if unparsed segments intervene.

(39)
| a | g | e | n | d | a |
|---|---|---|---|---|---|
|   | o |   | o | o |   |
| n |   | n |   |   | n |
|   |   |   | c |   |   |

The disposition of the consonant [n] depends on the ranking of the last two constraints. If NOCODA is higher ranked, 'c' is pruned and we get (incorrectly for English) the complex onset [nd]: *a.ge.nda.* If, on the other hand, *COMPLEX is higher ranked, then 'o' is pruned and [n] becomes a coda: *a.gen.da.*

The final module of the program massages representations like (40) into (41).

(40) a.  [a/[n],g/[o],e/[n],n/[o],d/[o],a/[n]]
     b.  [a/[n],g/[o],e/[n],n/[c],d/[o],a/[n]]

(41) a.  (a)(ge)(nda)
     b.  (a)(gen)(da)

**7.   The constraints**

In this section, I discuss certain interesting properties of the constraint system. Translating standard OT constraints into the theory proposed here results in some interesting asymmetries. To see this, let's adopt a convenient formalism. The following formal statement says that the structural option α is removed from elements of type X.

(42)  X
      |
      α

Some constraints are readily translated. For example, the NOCODA constraint simply removes the 'c' option from anything.

(43)  NOCODA    X
                |
                c

Other constraints are also easily recast in these terms, e.g. *PEAK/C, *MARGIN/V, and PARSE.

(44)  *PEAK/C       C
                    |
                    n

(45)  *MARGIN/V      V
                    |
                    {c,o}

(46)  PARSE         X
                    |
                    u

Other constraints in the implementation are more complex, e.g. *COMPLEX and ONSET. In the case of *COMPLEX, there are two issues. First, we need to penalize sequences of 'oo' or 'cc'. That is, *COMPLEX refers to complex margins, not complex onsets or codas. Any theory must make some sort of move here to generalize over the categories of onset and coda. (Interestingly, it is not altogether clear that we even want the theory to do this as there are plenty of languages that allow differing degrees of complexity in their onsets and codas.)

The other interesting issue facing a formalization of *COMPLEX is that it's symmetric. That is, focusing for the moment on onsets, we want *COMPLEX to apply in both of the following cases to remove the underlined candidates.

(47) 

| ... | C | C | ... |
|---|---|---|---|
|   | o | <u>o</u> |   |
|   |   | c |   |

| ... | C | C | ... |
|---|---|---|---|
|   | <u>o</u> | o |   |
|   | c |   |   |

This is easily done in the implementation, but requires a corresponding enrichment of the formalism. A constraint like (48) is interpreted as removing either α or β.

(48)  X Y
      | |
      α β

The *COMPLEX constraint can then be written as follows.

(49)  *COMPLEX      X X      X X
                   | |      | |
                   o o      c c

The ONSET constraint is similar in that it requires two clauses. The first clause penalizes any case of a segment preceding a 'n' that is specified anything but 'o'.

(50)  ONSET (clause a)    X   X
                          |   |
                        {n,c,u} n

The other clause penalizes anything specified 'n' in word-initial position.

(51)  ONSET (clause b)  [ X
                          |
                          n

These could easily be collapsed with an innovation in the formalism, but it is not clear that we should do so. There are plenty of languages with differing treatment of word-initial and medial onsetless syllables. These have usually been treated by appeal to some version of ALIGN-PRWD, but the current implementation suggests that we might be better served by exploring a different treatment based on splitting ONSET into two separate constraints.

Another issue raised by this implementation can be seen in the interaction between ONSET and *COMPLEX or NOCODA and *COMPLEX. The applicability of the *COMPLEX constraint to some segment depends on the values of a neighboring segment, so if those values haven't been set at the point *COMPLEX applies, *COMPLEX will skip over some cases. Consider a hypothetical form VCCVCC with *COMPLEX ranked above ONSET and NOCODA. For convenience, let's also assume that *PEAK/C, *MARGIN/V are all ranked above *COMPLEX., and that PARSE is ranked at the bottom. The candidate set starts out as follows.

(52)
| V | C | C | V | C | C |
|---|---|---|---|---|---|
| o | o | o | o | o | o |
| n | n | n | n | n | n |
| c | c | c | c | c | c |
| u | u | u | u | u | u |

After *PEAK/C, *MARGIN/V, and housekeeping, this is converted to the following.

(53)
| V | C | C | V | C | C |
|---|---|---|---|---|---|
|   | o | o |   | o |   |
| n |   |   | n |   |   |
|   | c | c |   | c | c |
| u | u | u | u | u | u |

Notice how if *COMPLEX is ranked at this point, it will not apply to either cluster. If NOCODA is ranked next, then, after PARSE gets its ultimate opportunity, a complex onset will surface in medial position, but unparsed consonants in final position.

(54)

| V | C | C | V | C | C |
|---|---|---|---|---|---|
|   | o | o |   |   |   |
| n |   |   | n |   |   |
|   |   |   |   |   |   |
|   |   |   |   | u | u |

The point is that the current algorithm implies that constraints are only relevant at one point in the CON-EVAL algorithm. This is different from what orthodox OT maintains.

There are two possibilities here. One would be to revise the algorithm to avoid this. This is readily done, but will obviously have a cost. The critical move is that after each constraint gets its opportunity, all higher-ranked constraints are reconsidered from top down.

The alternative would be to explore the consequences of this "one-shot" view of constraint application. The implications are obviously far-reaching, and an empirical investigation of them is beyond the scope of the current paper. I simply note that one-shot constraint application is more parsimonious in terms of complexity.

## 8. Alternative Models

In this section, I discuss other attempts at implementing OT and compare them to the current proposal. The particular proposals I consider are Black (1993), Ellison (1994), Hammond (1995), Tesar (1995), and Eisner (1995).

### 8.1. *Black (1993)*

Black (1993) offers the first attempt to implement OT computationally. He calls his approach "Constraint-Ranked Derivation". It's an interesting model, but is difficult to compare with the current proposal since it relaxes some key assumptions adopted in most of the OT literature.

First, Black proposes that rules create the candidates that constraints apply to. Second, though his constraints are ranked and violable, they are really only triggers for repair strategies. While the model avoids the problems of a potentially infinite candidate set, it does so by proposing a very different kind of OT.

### 8.2. *Ellison (1995)*

Ellison (1995) proposes that the output of GEN is a regular set and can be represented by a finite state automaton. Each constraint can be seen as a finite state transducer from the automaton produced by GEN to a list of harmony marks (constraint violations). Ellison proposes that constraints can be combined into a single transducer and the marks be organized into an ordered list. EVAL proceeds by pruning out suboptimal paths from the automaton.

This is an impressive effort and the proposal here can be seen as an elaboration of it. Basically, our proposal greatly restricts the class of automata GEN produces, and simplifies the CON-EVAL loop. Let's consider a simple example. An input like *hat* would get the following grid representation via GEN.

(55)

| h | a | t |
|---|---|---|
| o | o | o |
| n | n | n |
| c | c | c |
| u | u | u |

GEN can then be seen as a very simple finite state transducer from one string *hat* to another: oncu-oncu-oncu.

(56)
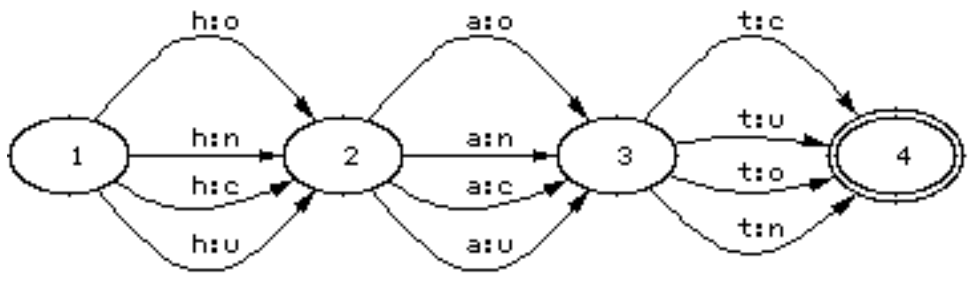

The constraints I've formalized above operate to prune arcs from these transducers.

The approach developed here then can be seen as a simplification of the proposal Ellison makes. The automata he proposes are greatly restricted and the transduction-pruning procedure he proposes is replaced with a simple set of pruning operations.[9]

*8.3. Hammond (1995)*

Hammond (1995) proposes an algorithm that is in many ways quite similar to the current one. It makes use of the same three central assumptions that the current proposal makes: i) no MAX or DEP violations are entertained, ii) there is a cyclic CON-EVAL pruning algorithm, and iii) candidates are encoded locally.

The differences between the current proposal and that earlier proposal are threefold. First, the earlier proposal was a very specific analysis of some facts in English and French and was intended to account for some very interesting facts in the psycholinguistic domain. The proposal did not offer a general account.

Second, the earlier implementation was coded in Perl, while the current implementation is in Prolog. The Prolog implementation offers a clear declarative semantics that permits more detailed analysis of the components of the program.

A third difference is that the '95 proposal maintained that a left-to-right segment-by-segment parse of the input was necessary to effect the local coding proposal. Moreover, if the rightmost segment was a consonant, then various constraints had to be skipped until the end of the word or a vowel was reached. The current proposal does not require such a parse, nor does it require a special stipulation for consonant-final strings.

---

[9] Given the nature of the transducers that GEN creates, it's quite easy to treat them as automata as well, where each symbol pair α:β is construed as a new symbol in an enlarged alphabet. On this view, the constraints of the current model can then be seen as transducers as well.

In fact, the current algorithm has been tested all-at-once, as illustrated above, and in a left-to-right fashion with the same special stipulation, and it performs equally well. Hence, left-to-right presentation is not essential to the current proposal, as it was to the '95 proposal.

*8.4. Tesar (1995)*

Tesar (1995) proposes an algorithm based on a technique called "dynamic programming". The basic idea is that the input is parsed segment by segment and that, at each point, only a finite number of candidate parses are maintained. As each segment is parsed, the best parse of the preceding string is determined and a new finite set of parses containing the preceding string and the current segment are then maintained. The following dynamic programming table for the input *at* can be used to exemplify the idea (p.14).[10]

(57)

|   | BOI | $i_1$=a | $i_2$=t |
|---|---|---|---|
| u |  | <a> | <at> |
| o | O | .Oa.O | .Oa.t |
| n | ON | Oa | .Oa.<t> |
| c | .ONC. | .OaC. | .Oat. |

At any point in the parse, the algorithm maintains four candidate outputs depending i) on the best parse of the preceding input, and ii) the different structural positions the current segment can fill. At the beginning of the input (BOI), four different candidates are maintained such that the last segment corresponds to essentially the same choices we have allowed. (Capital letters stand for an unfilled syllabic position: epenthesis.) When the first segment is encountered, four new candidates are selected based on i) what structural position the final segment (and only segment in this example here) occupies, and ii) how that is best achieved using the current constraint hierarchy and the four candidates from the previous stage (BOI here). The algorithm involves combining each of three basic operations (scan the segment, leave the segment unparsed, and epenthesize a segment) with the four preceding candidates and using the hierarchy to select a best candidate for whichever structural position one the current row commits one to.

The algorithm is a clever one and succeeds because of several assumptions that Tesar acknowledges. First, the number of rows must be finite. That is, only a finite number of instances of epenthesis are to be considered at each stage. As discussed above, epenthesis in natural language does not seem to be restricted in this way.

A second assumption is that only a single segment can occupy a structural position. This means, essentially, that complex onsets and codas are not considered in the candidate set. Tesar maintains that this can be done, but no specific proposal is worked out.

---

[10] I have changed Tesar's notation somewhat to bring it into conformity with that adopted in this paper.

A third assumption is that the constraints must be local. It must be possible to evaluate a constraint by looking at no more than two consecutive syllabic positions. As he acknowledges, constraints of the ALIGN family and various others fail to meet this criterion.[11]

*8.5. Eisner (1997)*

Eisner (1997) proposes a new formalization of OT that he terms "Primitive OT" (OTP). He formalizes an implementation of this theory in terms of the finite state methods developed by Ellison (discussed above). His proposal is relevant in the current context because he alters Ellison's proposal by adopting the equivalent of cyclic CON-EVAL pruning. As noted above, this results in an arithmetic savings over sequential CON-EVAL.

Eisner's proposal is restricted to the particular version of OT that he is developing. In addition to formalizing OT constraints in a restricted vocabulary, the model is also containment-based. Like Ellison's model, it does not treat the ALIGN family of constraints. The final section discusses Eisner's developing work on "factored automata".

*8.6. Summary of other models*

To summarize this section, the model developed here differs from previous models in a number of regards. The central proposals are, of course, as follows: i) MAX and DEP violations are not hypothesized, ii) cyclic CON-EVAL, and iii) local encoding of the candidate set.

In addition, the model exhibits a number of other properties which distinguish it from the other models discussed here. First, complex syllable margins can be treated. Second, there is no necessary locality of the constraints; constraints bearing on one element can refer to elements they are not adjacent to. Third, while the parser can function in a left-to-right fashion, it does not need such directionality to succeed. Fourth, Gen can be seen as a finite state transducer.

## 9. Conclusion

In this paper, I have proposed an implementation of Optimality Theory as a syllable parser. This parser is intended to model speakers' ability to judge the syllabic well-formedness of real and nonsense words.

The implementation made use of several innovations. First, I proposed that the parser does not hypothesize epenthesis or deletion. This makes sense given what the parser is modeling, but also alleviates the most glaring problem posed by implementing OT. The second innovation was cyclic CON-EVAL. This resulted in an arithmetic savings in computing time. The third innovation was local coding of the candidate set. This resulted in a exponential savings in computing time.

The implementation made a number of predictions. First, it suggests that lexical lookup should be best treated separately. (It leaves open the question of whether lookup should be modeled in terms of OT.) Second, it suggested a particular formalism for constraints. Third, it suggests that we should explore the implications of one-shot constraint ranking.

---

[11] Notice that the constraints we have proposed in our implementation satisfy Tesar's locality requirements, though this is not required by our implementation.

Michael Hammond
Department of Linguistics
University of Arizona
Tucson, AZ 85721

email: hammond@u.arizona.edu